# Anomaly detection optimization using big data and deep learning to reduce false-positive

Khloud Al Jallad[1*], Mohamad Aljnidi[1] and Mohammad Said Desouki[2]

*Correspondence: khloud.aljallad@hiast.edu.sy
[1] Faculty of Information Technology, Higher Institute for Applied Sciences and Technology, Damascus, Syria
Full list of author information is available at the end of the article

**Abstract**
Anomaly-based Intrusion Detection System (IDS) has been a hot research topic because of its ability to detect new threats rather than only memorized signatures threats of signature-based IDS. Especially after the availability of advanced technologies that increase the number of hacking tools and increase the risk impact of an attack. The problem of any anomaly-based model is its high false-positive rate. The high false-positive rate is the reason why anomaly IDS is not commonly applied in practice. Because anomaly-based models classify an unseen pattern as a threat where it may be normal but not included in the training dataset. This type of problem is called overfitting where the model is not able to generalize. Optimizing Anomaly-based models by having a big training dataset that includes all possible normal cases may be an optimal solution but could not be applied in practice. Although we can increase the number of training samples to include much more normal cases, still we need a model that has more ability to generalize. In this research paper, we propose applying deep model instead of traditional models because it has more ability to generalize. Thus, we will obtain less false-positive by using big data and deep model. We made a comparison between machine learning and deep learning algorithms in the optimization of anomaly-based IDS by decreasing the false-positive rate. We did an experiment on the NSL-KDD benchmark and compared our results with one of the best used classifiers in traditional learning in IDS optimization. The experiment shows 10% lower false-positive by using deep learning instead of traditional learning.

**Keywords:** Intrusion detection systems (IDS), Security intelligence optimization, Unknown threats, Big data, NSL-KDD dataset, False-positive

## Introduction

Signature-based Intrusion detection systems are not suitable anymore to be used in nowadays network environment. Because signature-based models are not able to detect new threats and unknown attacks. Due to technology improvement, the number of attacks is increasing exponentially. Statistics show that attacks number increases with a rate of 100% each year causing huge money loss, about tens of millions of dollars for ransomware attacks only. This high number of millions of new threats that are developed every day, reduces the effectiveness of signature-based IDS because it is not a practical solution to update the signatures databases every few minutes. Anomaly-based IDS can be a better alternative of signature-based IDS because it is more suitable for nowadays challenges of new threats. Because it can detect new threats but still not practically used because of its high false-positive rate. The reason behind the high false-positive rate is that anomaly model classifies an unseen pattern that did not learn it in normal cases, as an abnormal case. The reason behind high false-positive is that the model has overfitting what means it cannot generalize. The solution may seem easy, by extending the training dataset to include all normal cases. But that still not a practical solution for a long time. Although we can add much more normal cases to datasets, we still need a model with higher ability of generalization. In this paper, we propose using deep learning with big data to solve this problem. Big data allows us to use big datasets for training to reduce the false-positive rate by including much more normal cases. And deep learning needs large datasets for training without facing the overfitting problem as it may have more ability to generalize than traditional learning models. We compare results with one of the best traditional learning used classifiers on the NSL-KDD benchmark dataset. Results show reducing false-positive to lower 10% than already used classifiers.

The paper is organized as follows, we will talk about related works in "Related Work Section". The proposed method is explained in detail in "Method Section". "Data Section" contains detailed information about used dataset. We will discuss results in "Results and Discussion Section". We will talk about conclusion and future vision in "Conclusion and Future Work Section".

## Related work

To optimize IDS, there are two general resolving directions. First way is to collect more data. Network security data and intrusion data are hard to collect because of data privacy concerns. In addition to the challenges of labeling data, which may be time-consuming for experts to do explanations and labeling process. Second way is studying how to increase the performance on small datasets, which is very important because the insights we can get from such researches can be implemented in big data researches. This paper is considered under the second way. We propose implementing deep learning instead of traditional learning. We choose to compare our works with SVM, shown in Fig. 1. Because SVM is the most popular traditional learning model used in network security and intrusion detection systems along years even in the era of big data. SVM has received a lot of interest in IDS optimization domain because it has proved by lots

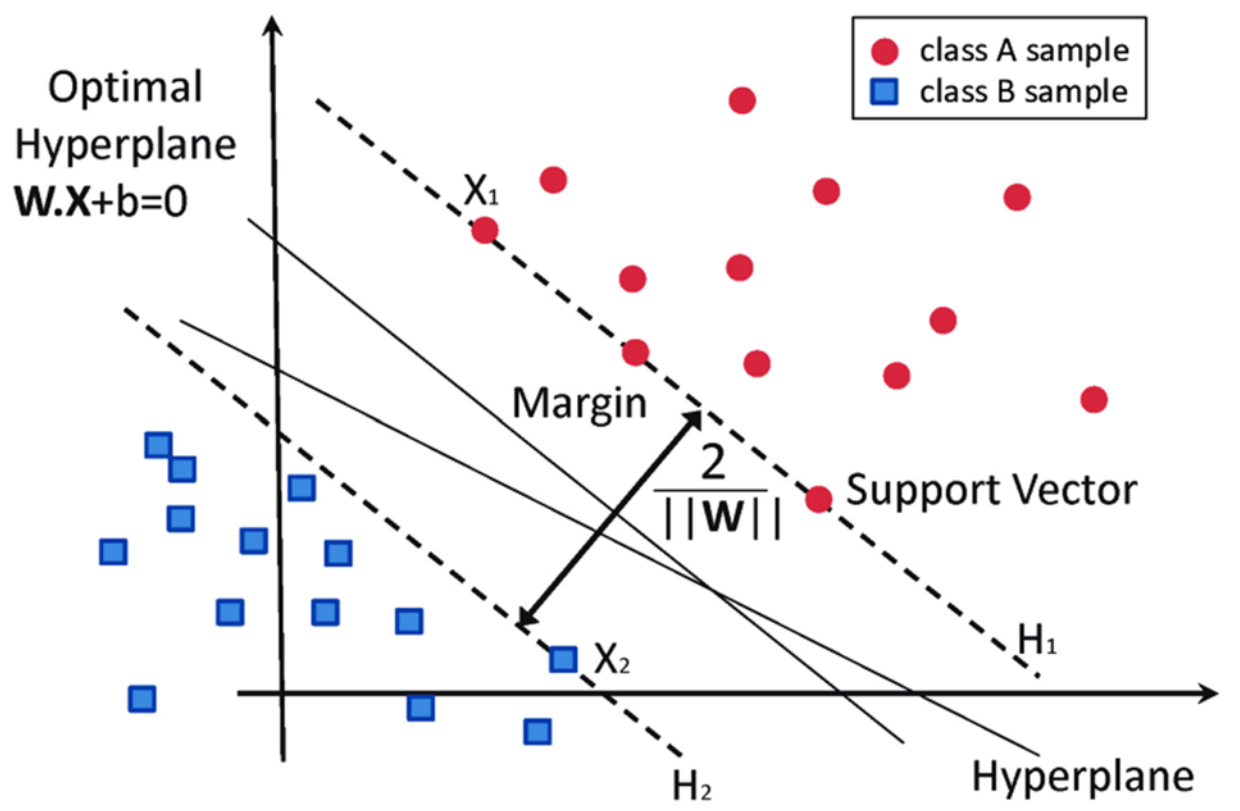

**Fig. 1** SVM [6]. Shows the main concept of SVM. Its margins and support vectors

of experiments that it is one of the best classifiers in anomaly-based intrusion detection systems [1–5]

Find Table 1 from [1], that contains research experiments that used basically SVM and optimized it by adding another model to its result or to its input.

## Method

The problem of any anomaly model is high false-positive rate, because it classifies any unseen pattern as abnormal where it may be a normal pattern but not included in the training set. Having a big training dataset that includes all possible normal cases has been considered as a theoretical idea that could not be applied in practice. Although we can increase the number of training samples to include much more normal cases, still we need a model that has more ability to generalize. We propose deep learning model instead of traditional learning as it may have more ability to generalize.

The proposed solution is to use a big data training set, but also to use a deep learning model to solve overfitting problem that occurs on the traditional learning models.

We have done many experiments [26] that we care about collective and contextual attacks. Where we chose using a deep network with type of recurrent so we can handle sequences of actions or events. We chose LSTM to deal with sequences of inputs and focus on context.

We were curious if LSTM can also achieve better results on small benchmark datasets or not. So we applied this experiment, we do not use sequences or even big data in this experiment. Only small benchmark data is used with LSTM.

### Proposed algorithm-long short term memory(LSTM)

LSTM is type of Deep Learning Recurrent Neural Network (RNN). see Fig. 2. RNN is extension of a convention feed-forward neural network. Unlike feedforward neural networks, RNN have cyclic connections making them powerful for modeling sequences. As a human no one think of each event separately. For example, when you are reading this article you read each word but you understand it in the context of this article, so that you understand the whole concept of this paper. That is the idea of RNN that has a loop to deal with input as a sequence, and that what we need to handle each event on network within its context.

LSTM is a special case of RNN that solves problems faced by the RNN model [27, 28]

1. Long term dependency problem in RNNs.
2. Vanishing Gradient & Exploding Gradient

LSTM is designed to overcome vanishing gradient descent because it avoids long-term dependency problem. To remember information for long periods of time, each common hidden node is replaced by LSTM cell. Each LSTM cell consists of three main gates such as input gate *it*, forget gate *ft*, and output gate *ot*. Besides *ct* is cell state at time *t*.

LSTM architecture is shown in Fig. 3, LSTM cell is shown in Fig. 4.

Figure 5 shows the equations to calculate the values of gates

**Table 1 Comparisons between Researches used SVM as basic Classifier**

| Approach | Working idea | Dataset | Detection rate | False alarm |
|---|---|---|---|---|
| Unsupervised anomaly detection system [7] | Tune and optimize automatically the values of parameters without pre-defining them | From Kyoto University honeypot | – | – |
| Multiclass SVM [8] | Attributes are optimized using k-fold cross validation. This technique can be used to decrease the rate of False-Negatives in the IDS | Self | – | – |
| OC-SVM One-Class SVM [9] | Multistage OC-SVM and feature extraction represents a method to detect unknown attacks. Method is poor in second stage classifier to detection rate of unknown attacks | From Kyoto | 80.00 | 20.94 |
| IG-ABC-SVM Information Gain-Artificial Bee Colony [10] | A combining IG feature selection and SVM classifier in IDS model is proposed. Experiments using just two swarm intelligence algorithms | NSL-KDD | 98.53 | 0.03 |
| SbSVM [11] | Autonomous labeling algorithm of normal traffic (when the class distribution is not imbalanced) Not evaluated for real time case | DARPA | 99 | 5.5 |
| RS-ISVM- reserved set -Incremental SVM [12] | An incremental SVM training algorithms is used, hybrid with modifying kernel function U-RBF Foreseeing attacks, specifically for attacks of U2R and R2L may not tolerate but oscillation problem solved | KDD Cup 1999 | 89.17 | 4.9 |
| SVM-GA [13] | Hybrid model by combining (GA and SVM) | KDD CUP 1999 | 98.33 | 0.50 |
| Genetic principal Component [14] | Subset selection using GA and PCA | KDD cup 1999 | 99.96 | 0.49 |
| SVM and NN [15] | Hybrid process Most significant performance as far as training time but time consuming and hard task to trigger | DARPA | 99.87 | – |
| N-KPCA-GA-SVM kernel principal component analysis-genetic algorithm (GA)-SVM [16] | Hybrid of KPCA, SVM and GA algorithms. Faster convergence speed. Performs higher predictive accuracy and better generalization But have complex structure and have latency for real time application | KDD CUP99 | 96.37 | 0.95 |
| CSV-ISVM Candidate Support Vector -Incremental SVM [17] | Candidate Support Vector -Incremental SVM improves detection rate and false alarm rate | KDD Cup 1999 | 90.14 | 2.31 |
| Hybrid approach of K-Medoids, SVM and Naïve Bayes [18] | Hybrid learning approach through a combination of K-Medoids clustering, Selecting Feature using SVM, and Naïve Bayes classifier | KDD Cup 1999 | 90.1 | 6.36 |

**Table 1 (continued)**

| Approach | Working idea | Dataset | Detection rate | False alarm |
|---|---|---|---|---|
| Distance of sum-based SVM [19] | Hybrid learning method named distance sum-based support vector machine (DSSVM) | KDD Cup 1999 | | |
| SVM and GA [20] | Hybrid method consisting of GA and SVM for intrusion detection system | KDD Cup 1999 | 0.98 | 0. 017 |
| PCA and SVM [21] | Hybrid model by integrating the principal component analysis (PCA) and (SVM) | NSL-KDD | 0.9655 | – |
| SVM with GA [5] | FWP-SVM-genetic algorithm (feature selection, weight, and parameter optimization of support vector machine based on the genetic algorithm) | KDD Cup 1999 | 96.61 | 3.39 |
| SVM for Anomaly in smart city wireless network [22] | Compare SVM and isolation forests to detect anomalies in a laboratory that reproduces a real smart city use case with heterogeneous devices, algorithms, protocols, and network configurations | smart city WSNs | 85 | 5%-10% |
| SVM and RBF and [23] | SVM using Radial-basis kernel function (RBF) and a Particle Swarm Optimization algorithm to optimize the parameters of SVM | NSL-KDD | 99.8 | 0.9 |
| SVM and GA [24] | Hybrid classification algorithm (GSVM) based Gravitational Search Algorithm (GSA) and support vector machines (SVM) to optimize the accuracy of the SVM classifier by detecting the subset of the best values of the kernel parameters for the SVM classifier | KDD CUP 99 | 97.5 | 0.03 |
| Support Vector Machine (SVM) Based on Wavelet Transform (WT) [25] | Support Vector Machine SVM based on Wavelet Transform(WT) | UNSW-NB15 | 95.92 | – |

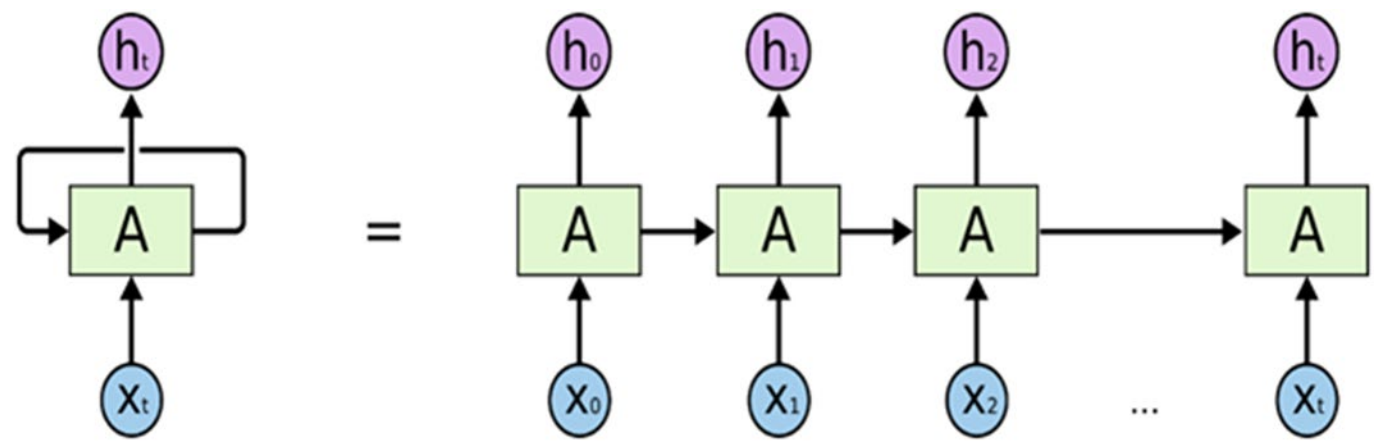

**Fig. 2** RNN have Loops [27]. Shows simple diagram of recurrent neural networks

where *xt*, ht, and *ct* correspond to input layer, hidden layer, and cell state at time *t*. Besides, *bi*, *bf*, *bc*, and *bo* are bias at input gate, forget gate, cell state, and output gate, respectively. Furthermore, σ is sigmoid function. Finally, *W* is denoted by weight matrix

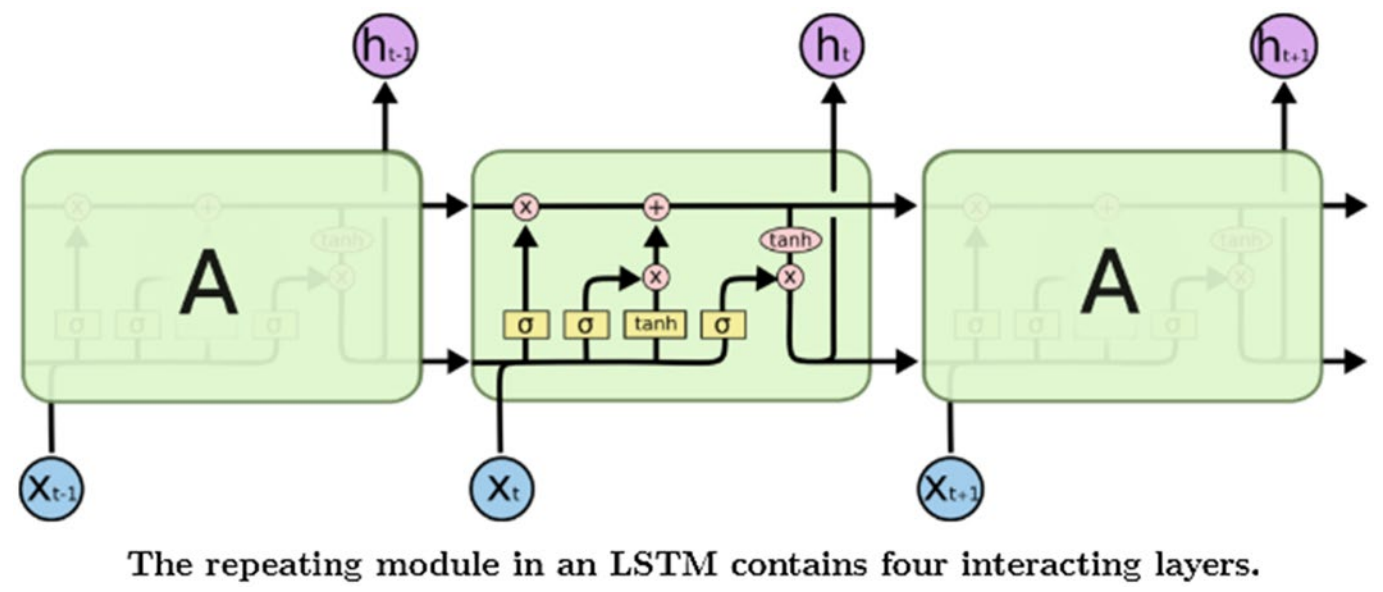

**Fig. 3** LSTM architecture [39]. Shows in details architecture of long short term memory

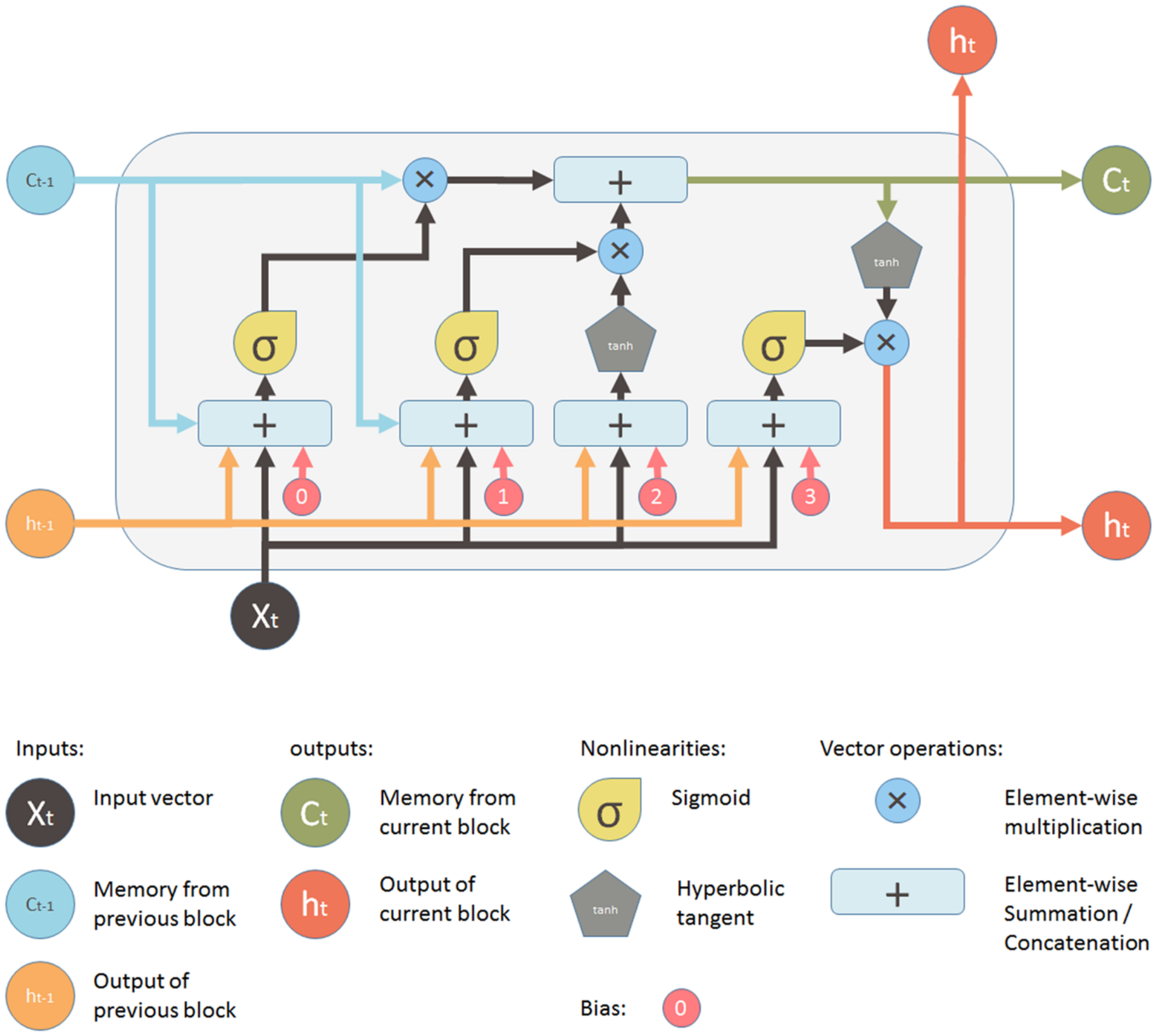

**Fig. 4** LSTM cell [29]. Shows in details architecture of long short term memory

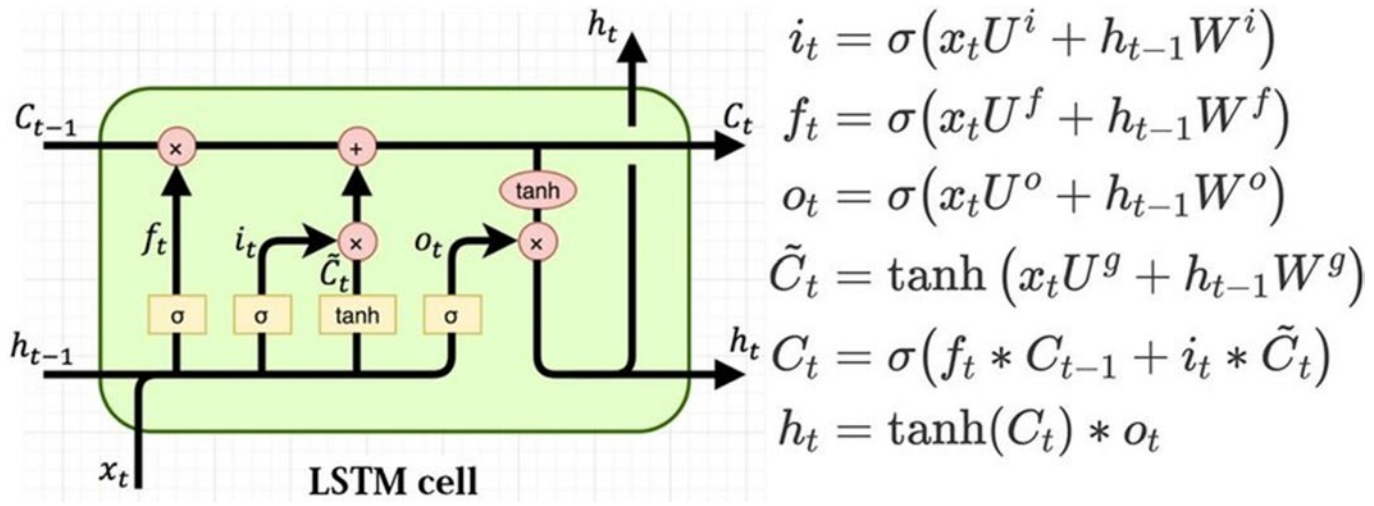

**Fig. 5** LSTM cell equations [38]. Shows in details equations of LSTM Cell

**Table 2 Statistics about KDD training set [33]**

| | Number of samples | Number of distinct samples | Possible reduction percentage |
|---|---|---|---|
| Attacks | 3,925,650 | 262,178 | 93.32% |
| Normal | 972,781 | 812,814 | 16.44% |
| Total | 4,898,431 | 1,074,992 | 78.05% |

**Table 3 Statistics about KDD testing set [33]**

| | Number of samples | Number of distinct samples | Possible reduction percentage |
|---|---|---|---|
| Attacks | 250,436 | 29,378 | 88.26% |
| Normal | 60,591 | 47,911 | 20.92% |
| Total | 311,027 | 77,289 | 75.15% |

### Experiment setup

This experiment is performed on Google CoLab [30] using Keras libary (Python Deep Learning library) [31]. We apply LSTM of 64 hidden nodes with Relu activation function and dropout = 0.5. Using binary cross-entropy loss function. Using RMSprop optimizer. Learning rate = 0.001, rho = 0.9, decay = 0.0.

## Data

### KDD99

This dataset is an updated version of the DARPA98 by processing the tcpdump portion. it was constructed in 1999 by the international competition, International Knowledge Discovery and Data Mining Tools Competition. Its size is 708 MB and it contains about 5 million connections [32]. It contains different attacks such as Neptune-DoS, pod-DoS, SmurfDoS, and buffer-overflow. The benign and attack traffic are merged together in a simulated environment but it contains a large number of redundant records [33]

KDD99 is the most famous research dataset but it cannot be used for real-life applications as data is old, not real, not enough. Statistics details are provided in Table 2 and Table 3.

#### *Statistics about KDD training set*

#### *Statistics about KDD testing set*

### NSL-KDD

This dataset is available online on website of Canadian Research for Intrusion detection [34]

NSL-KDD is built by resampling KDD99 to solve all its drawbacks that reduce the precision of mining algorithms on it. Resampling is done by removing duplicated samples and redundant samples. As KDD training set has 78% of duplicated samples and testing

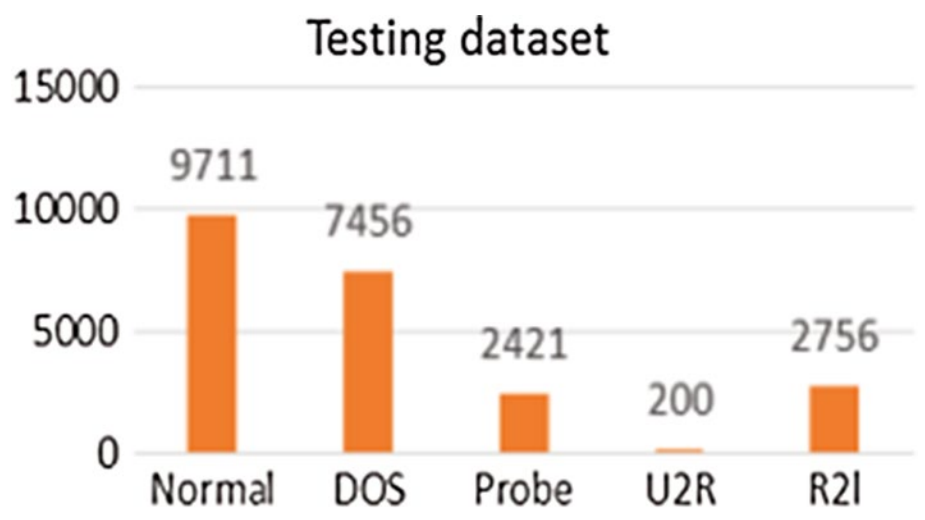

**Fig. 6** Statistics about NSL-KDD testing dataset [3]. Shows Statistics about NSL-KDD training datasets. Statistics about number of samples for each type of attack

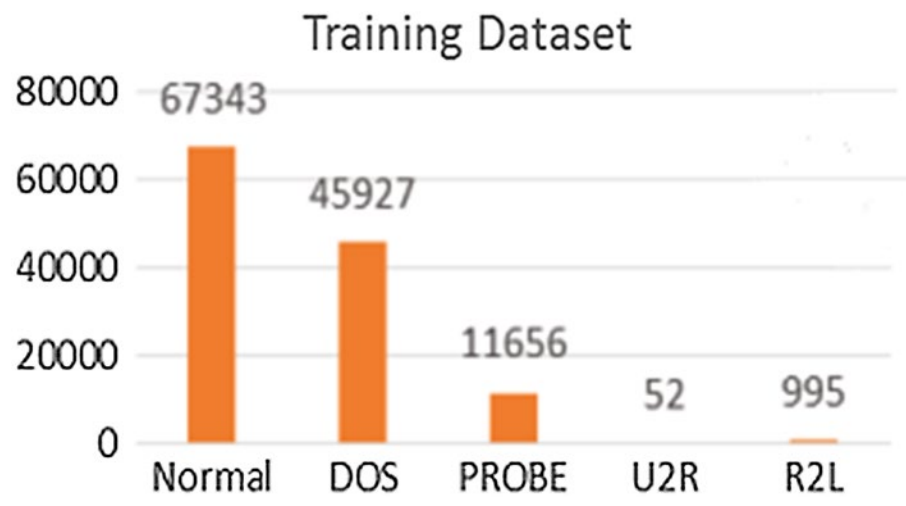

**Fig. 7** Statistics about NSL-KDD training dataset [3]. Shows Statistics about NSL-KDD testing datasets. Statistics about number of samples for each type of attack

set has 75% of duplicated samples that result to bias detection to detect attacks that have more number of samples and make it hard to learn attacks that have small number of samples but may be one most dangerous attacks, such as U2R, L2R.

### *NSL _KDD advantages*

- It does not contain redundant samples in testing samples so that solves bias problem
- It does not contain duplicate records in the test set which have better reduction rates.
- All samples for any attack type has the same percentage of its number in KDD.
- It contains 21 attacks in training dataset and 37 attacks in testing dataset so we can test the unknown attacks detection capability of a model by testing on attacks that some of them are new for the model as they were not in training set [3, 33] .

### *NSL statistics*

Statistics about NSL_KDD training and testing datasets in Figs. 6, 7.

Where x-axes represent attacks types, y-axes represent number of samples.

NSL-KDD is not biased but that does not mean that it is optimal. It has many drawbacks, but still more effective than other available datasets. At least, it will become out-of-date after few months as developing technologies and new attacks appear every day.

**Table 4 Type of attacks available in NSL-KDD [3]**

| Attack type | Attack name |
|---|---|
| DoS | Back, Land, Neptune, Pod, Smurf, Teardrop, Mailbomb, Processtable, Udpstorm, Apache2, Worm |
| R2L | Guess_password, Ftp_write, Imap, Phf, Multihop, Warezmaster, Xlock, Xsnoop, Snmpguess, Snmpgetattack, Httptunnel, Sendmail, Named |
| U2R | Buffer_overflow, Loadmodule, Rootkit, Perl, Sqlattack, Xterm, Ps |
| Probe | Satan, IPsweep, Nmap, Portsweep, Mscan, Saint |
| Normal | Normal |
| Unknown | Unknown |

| F# | Feature name | F# | Feature name | F# | Feature name |
|---|---|---|---|---|---|
| F1 | Duration | F15 | Su attempted | F29 | Same srv rate |
| F2 | Protocol type | F16 | Num root | F30 | Diff srv rate |
| F3 | Service | F17 | Num file creations | F31 | Srv diff host rate |
| F4 | Flag | F18 | Num shells | F32 | Dst host count |
| F5 | Source bytes | F19 | Num access files | F33 | Dst host srv count |
| F6 | Destination bytes | F20 | Num outbound cmds | F34 | Dst host same srv rate |
| F7 | Land | F21 | Is host login | F35 | Dst host diff srv rate |
| F8 | Wrong fragment | F22 | Is guest login | F36 | Dst host same src port rate |
| F9 | Urgent | F23 | Count | F37 | Dst host srv diff host rate |
| F10 | Hot | F24 | Srv count | F38 | Dst host serror rate |
| F11 | Number failed logins | F25 | Serror rate | F39 | Dst host srv serror rate |
| F12 | Logged in | F26 | Srv serror rate | F40 | Dst host rerror rate |
| F13 | Num compromised | F27 | Rerror rate | F41 | Dst host srv rerror rate |
| F14 | Root shell | F28 | Srv rerror rate | F42 | Class label |

**Fig. 8** NSL KDD features [34]. Shows features of NSL-KDD dataset

### *Type of attacks available in NSL-KDD*

NSL-KDD contains 37 types of attacks that are categorized into 6 basic categories.

Categories are DoS, R2L, U2R, Probe, Normal, Unknown. Find details in Table 4.

### *NSL-KDD attributes*

NSL-KDD contains 41 attributes as we can find in Fig. 8 [35].

## Results and discussion

We will not discuss that big data is better with deep learning than traditional learning. It is a fact that the most important difference between deep learning and traditional machine learning is its performance as the scale of data increases. As we can see in Fig. 9 [36].

We did not use a big dataset in this experiment. We used a small dataset as we want to compare the results of applying deep learning instead of traditional learning in anomaly-based IDS even on small dataset. Thus, we can compare optimization of generalization. We got by experiment, that accuracy increases by using deep learning model instead of traditional learning model. Moreover, false positive is getting lower by 10% less than traditional learning models. Results are shown in Table 5.

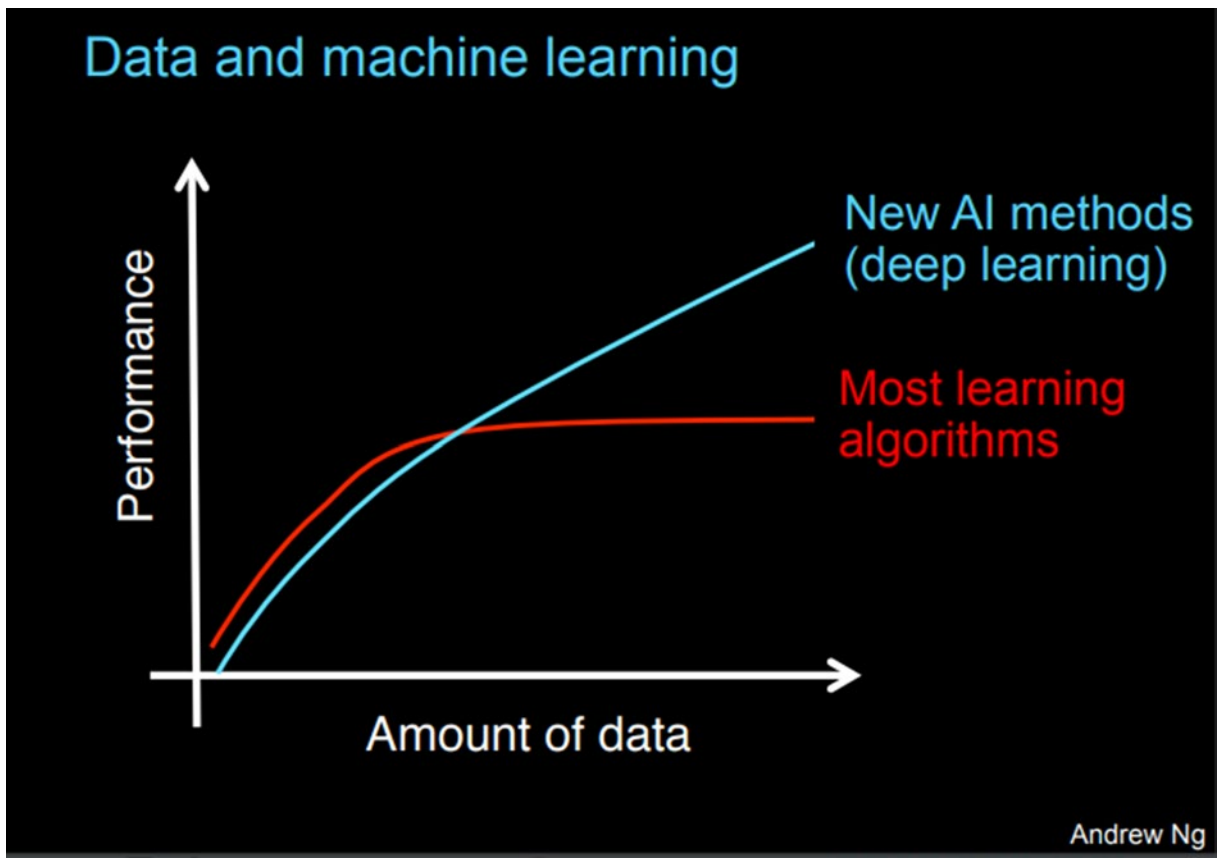

**Fig. 9** Big data with deep learning performance optimization [36, 37]. Shows performance optimization when using bigger amount of data

**Table 5 Comparing false-positive of Deep vs SVM on NSL-KDD Benchmark**

| Methodology | False-positive FP | False negative FN | Accuracy |
|---|---|---|---|
| Deep learning (LSTM) | 0.01 | 0.03 | 0.9676 |
| SVM | 0.1 | 0.03 | 0.87 |

We used SVM to compare with, because it has one of the highest results among traditional learning classifiers. The result we get is higher than all previous studies we have shown in related works.

## Conclusion and future work

Using big data analysis with deep learning in anomaly detection shows excellent combination that may be optimal solution as deep learning needs millions of samples in dataset and that what big data handle and what we need to construct big model of normal behavior that reduce false-positive rate to be better than small traditional anomaly models.

We recommend using deep learning models instead of SVM when trying to build hybrid classifiers for IDS. Because, as we see in results, deep models have more ability to generalize than traditional models. Thus, deep models achieve better results on unseen data causing less false-positive rate. We recommend using LSTM as we think it is an optimal solution in security domain because it can deal with sequences of events and context. So that it cannot only achieve better accuracy, but also can detect more various types of attacks that were not possible before. Such as, contextual and collective attacks.

### Abbreviations

IDS: Intrusion detection system; RNN: Recurrent neural networks; LSTM: Long short term memory; FP: False-positive; DoS: Deny of service attack; R2L: Remote to user; U2R: User to root.

**Acknowledgements**
I would like to thank my parents for their endless love and support.

**Authors' contributions**
KHJ took on the main role so she performed the literature review, conducted the experiments and wrote the manuscript. MJ and SD took on a supervisory role and oversaw the completion of the work. All authors read and approved the final manuscript.

**Funding**
The authors declare that they have no funding.

**Availability of data and materials**
All datasets in this survey are available online, you can find links in references.

**Ethics approval and consent to participate**
The authors Ethics approval and consent to participate.

**Consent for publication**
The authors consent for publication.

**Competing interests**
The authors declare that they have no competing interests.

**Author details**
[1] Faculty of Information Technology, Higher Institute for Applied Sciences and Technology, Damascus, Syria. [2] Faculty of Information Technology, Arab International University, Daraa, Syria.

## Publisher's Note